# Evaluating Usage of Images for App Classification

Kushal Singla, Niloy Mukherjee, Hari Manassery Koduvely, Joy Bose
Samsung R&D Institute
Bangalore, India
kushal.s@samsung.com

*Abstract*— App classification is useful in a number of applications such as adding apps to an app store or building a user model based on the installed apps. Presently there are a number of existing methods to classify apps based on a given taxonomy on the basis of their text metadata. However, text based methods for app classification may not work in all cases, such as when the text descriptions are in a different language, or missing, or inadequate to classify the app. One solution in such cases is to utilize the app images to supplement the text description. In this paper, we evaluate a number of approaches in which app images can be used to classify the apps. In one approach, we use Optical character recognition (OCR) to extract text from images, which is then used to supplement the text description of the app. In another, we use pic2vec to convert the app images into vectors, then train an SVM to classify the vectors to the correct app label. In another, we use the captionbot.ai tool to generate natural language descriptions from the app images. Finally, we use a method to detect and label objects in the app images and use a voting technique to determine the category of the app based on all the images. We compare the performance of our image-based techniques to classify a number of apps in our dataset. We use a text based SVM app classifier as our base and obtained an improved classification accuracy of 96% for some classes when app images are added.

## I. INTRODUCTION

App classification refers to the problem of classifying mobile apps into one of a set of categories defined as per a given taxonomy. It may be done manually or by using tools such as machine learning techniques. There are a number of such tools available with varying degrees of accuracy. However, most of the existing tools rely on text descriptions of the apps in order to classify them. Such text based approaches may not be useful in certain cases such as when the amount of text in the app description is wrong, too small or does not describe the app adequately. Fig. 1 shows screenshots of a few apps where the text descriptions are too short.

In such cases, one solution can be to use the images associated with the app to complement the textual information, and thus increase the accuracy of the classification. We assume here that app images might be more descriptive of the actual content of the app in some cases as least and give a higher precision when used in classification of the app.

In this paper, we seek to evaluate different methods in which app images can be used to improve the accuracy of the app classification. One such method involves extracting text from the app images using optical character recognition (OCR) and using the extracted text to classify the app. Another method involves generating text descriptions of the app images by summarizing the images using a tool, and using the resulting text descriptions for the app classification. Yet another method involves identifying the objects in the app images and using the identified objects to classify the app. An ensemble of such different models can also be used, perhaps along with text based classification of apps. We hope that our exploration in this paper would encourage users to use app images as an additional factor when determining the classification category of a given app.

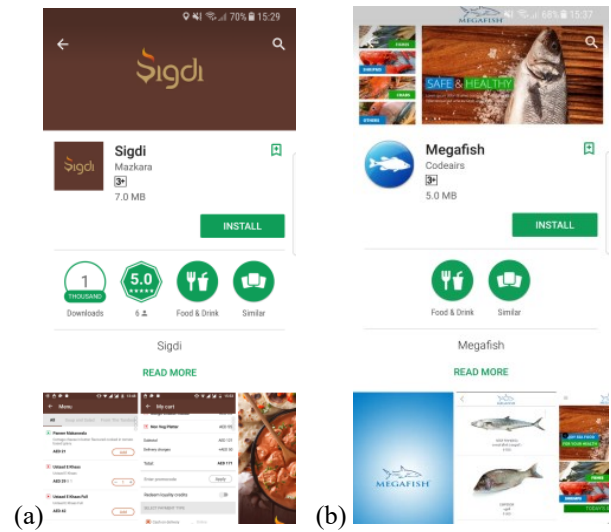

Fig. 1 Screenshots of mobile apps related to food category, where the text descriptions are too short or inadequate.

The rest of the paper is organized as follows: first we consider related work in the fields of app classification and image identification in section 2. Then we describe each of our approaches along with our evaluation methodology for that approach in section 3. Section 4 gives the experimental setup and our classification results on a dataset of apps. Section 5 concludes the paper.

## II. RELATED WORK

There are a number of works related to the field of automatic app classification. Zhu et. al. [1, 2] have created app taxonomies based on the real world context learnt from the user's online behavior, such as URLs browsed or searches in a search engine. A similar approach was employed by Shevale [3]. Lindorfer [4] also built a system to classify apps, although their motivation was to analyze unknown apps for security considerations such as risk of malware. Padmakar [5] similarly calculated a risk score for unknown apps. Seneviratne [6] developed a system to identify spam apps using app metadata. Olabenjo [7] used a Naïve Bayes model to automatically classify apps on the Google play store based on app metadata, using existing app categories to learn categories of new apps. Radosavljevic [8] used a method to classify unlabeled apps using a neural model trained on smartphone app logs.

Object detection is a computer vision technique which uses convolutional neural networks to identify the semantic objects in an image. Usilin [9] discussed the application of object detection for image classification. One of the approaches described in this paper uses an object detection method, but for app classification rather than image classification.

There are also a number of works using images for document classification. Junker [10] evaluated various techniques using Optical Character Recognition (OCR) extracted text for document classification. More recently, Bouguelia [11] used OCR to extract text from images, using the extracted text to classify documents in their dataset using a number of methods including Naïve Bayes and active learning.

Another way to use images for classification is to generate image captions, on which the text classification algorithms are run. Convnets and recurrent nets based methods to create image captions are described in several papers, such as by Karpathy et al [12]. Gan [13] used a system to generate image captions using CNNs and RNNs.

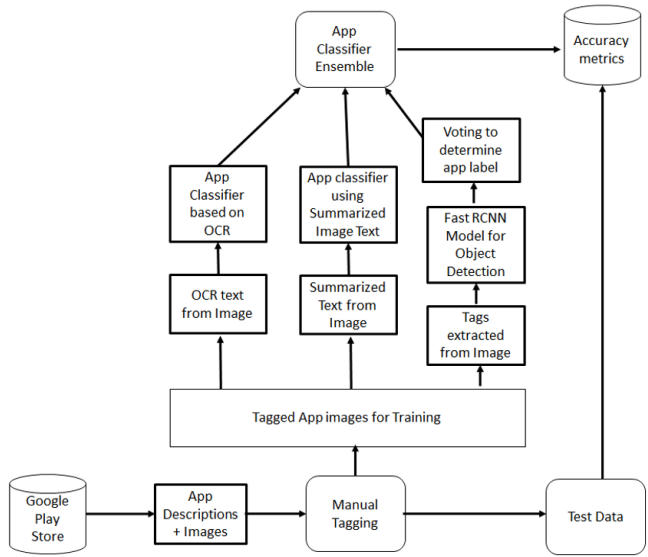

Fig. 2. High level architecture for app image based app classification.

### III. APPROACHES TO IMAGE BASED APP CLASSIFICATION

In this section we describe a few different approaches that can be used for app images based app classification. All these approaches can be evaluated either individually or as an ensemble, either purely based on app images or on a combination of text metadata for the app and the app images.

In our system, the steps to be followed are: a number of app images along with app descriptions are extracted from an app store such as the Google play store, and manually tagged with the app category labels from a fixed number of app categories. The apps are divided into a training set and a test set. After this, the app images are processed and text extracted using the various methods described below. The extracted text from the app images in the training set is used to train the classification models for each method, and the same models are used on the test set to test the classification accuracy.

Fig. 2 shows the high level architecture of our system.

#### A. Optical character recognition (OCR) based method

In the OCR method, first we extract text from app images via OCR for each of the apps in the training set. Any good open source OCR library can be used for our purpose. The extracted text from all the app images is then used to train the classification model, using SVM.

The advantage of the OCR based approach lies in its simplicity and ease of use. Often, the OCR data contains useful information about the app image that can help to further refine the app classification accuracy. Approaches that do not use OCR miss out on this additional data. However, one problem with the OCR based approach is that the quality of text extracted is often not very high, since there is too much noise in the text. So as a standalone approach it may not be useful, but in combination with other approaches might enhance their value.

#### B. Image summarization based method

In this method, we use image summarization to create a description of each of the app images. The summary can be a few short sentences describing the image, such as what are the objects within the image and the relationship between these objects. An example summary can be 'cat standing on a table.' We then train the app classifier to fit the app image summary to the app classification label, using the extracted text from the app image summaries.

A number of methods for image summarization or scene summarization are available in the literature [14-16]. Any of the existing methods or tools can be used to generate captions or text summaries for the app images.

The advantage of the image summarization based approach for app classification is that the summarized app images may contain useful data about the usage of the app, such as the app screens and so on. Moreover, since summarization contains the relation between different objects identified in the app image, it can be useful in any language. Especially in cases where the app description is too small or in a different language, the app summarization method can be useful.

#### C. Convolutional neural network based object detection method

In this method, we use a convolutional neural network to detect objects within the app images. The bounding boxes for unique objects in each app image in the training set are first manually drawn and tagged with object labels. The images with bounding boxes with object labels are used to train the fast RCNN classifier [17, 18] to detect object labels within the app images in the test set. At the time of inferencing, for each app, we get all the images of the app and label the objects detected in the image.

Some advantages of using neural networks for object detection is the speed of object detection, once trained it can be performed in real time. Moreover, such models can be trained on less data as well. One disadvantage is that the number of training cycles can be more, also as the number of classes increases, the model size and training time will also go up.

### D. Pic2Vec based method

In this approach, we use the pic2vec model [19], which is trained on an images dataset to create an embeddings vector for each image, to convert an app image into a real valued vector.

We first generate vectors for each of the app images, then take the average of the vectors for all the app images. After this, we train an SVM model to associate the vector to the app label. Once the model is trained on multiple apps in our training set, we then use it in our test set to classify the apps in our test set.

The advantages of the embeddings based methods like Pic2Vec lie in the translation invariance, intuitiveness and maintenance of relations of concepts between similar images. Disadvantages include the need for huge training data.

To summarize, the different methods for app classification using app images all have their advantages and disadvantages. We evaluate the methods individually and in combination as an ensemble.

---

ALGORITHM: Ensemble based ranking method to determine the final app label

---

1. Input:
$A_I$ = Apps for inferencing where $1<=I<=n$
$I_{IK}$ = $K^{th}$ image for $I^{th}$ App where $1<=K<=m$
$B_{IKJ}$ = $J^{th}$ bounding box for the $K^{th}$ image of the $I^{th}$ App
$L_{IKJ}$ = Label for the $J^{th}$ bounding box for the $K^{th}$ image of the $I^{th}$ App
2. Output:
$A_I^L$ = App label for the app where $1<=I<=n$
$A_{IK}^L$ = Label of the Kth image of the app where $1<=K<=m$
3. For each App $A_I$
4.    For each Image $I_{IK}$
5.       For each bounding box $B_{IKJ}$
6.          Fetch the label $L_{IKJ}$
7.    Label the image($A_{IK}^L$) with the label of the maximum bounding boxes.
8.    Label($A_I^L$) the app($A_I$) with the label of max number of images.
9. *exit*: end procedure

---

### E. Ensemble method to calculate the final app label using Borda count

An ensemble method of combining the individual classification methods is by using a modified version of the Borda count method [20], involving taking a vote of each app image as to which class it belongs to, and taking the majority of the votes from the app images as the classification of the app.

In the following section, we describe the experimental setup and results of our model to measure the classification accuracy of each image based method for app classification.

## IV. EVALUATION OF IMAGE BASED APP CLASSIFICATION METHODS

### A. Tool details

For the OCR method, we have used a crawling tool to crawl the app database from Google play. We used pytasseract as the OCR tool. Pytesseract [21] is a python library for extracting text from images, based on the Tasseract open source OCR engine [22]. It uses the Google OCR Engine with various models including LSTM.

For image summarization, we used an online Microsoft tool called captionbot [23] for our purpose of getting the summarized image descriptions. Captionbot [23] allows the user to upload an image and returns a description of the image in one or more sentences, along with a degree of confidence.

For the convolutional neural net approach, we used the Microsoft Visual Object Tagging Tool (VOTT) [24] for tagging and identifying the object boundaries. VOTT uses one recurrent neural net to identify the boundaries and another to tag. We also used a fastRCNN classifier [17, 18] to detect object labels within the app images in the test set.

Presently, due to lack of time we did not evaluate the Pic2Vec method for image based app classification. However, this can also present an additional way to classify apps and can be included in an ensemble with the other models.

### B. Experimental setup

For our experiment to measure the accuracy of the app classification using the image based methods, we use a dataset of app descriptions and the associated app images along with the app category labels. We choose apps from the following three categories: food, beauty and décor.

For validation, we have 695 apps labelled into one of these categories: beauty, food, décor. We have not balanced the sampling of apps here, so our choice of apps reflects the distribution of apps in the app store. Out of this, we choose 330 as the training set (ground truth) and 365 as the test set. We train the model to classify the apps in the training set from the app images using three methods described in the previous section: OCR, image summarization and RCNN based object detection, along with the text based app classification method as a baseline.

We used an SVM model with a linear kernel with the following parameters: C = 1.0, gamma = 0.001. We performed feature selection using Chi square to select the top 100 features. For text data we used TF-IDF encoding.

After getting the individual app classification results in our test set using OCR and other models, we use an ensemble model using a voting system to get the resultant category of each app from its images.

The ensemble voting system works as follows: Each of the classification methods (3 image based, and one text based) votes for a category for the app. The majority vote is the resultant category. In case of a tie in voting, the category predicted by the text based method is given preference.

C. *Classification results*

The classification accuracy (F1 score) using the different methods is shown in table 1, along with a graph plotted in Fig. 3.

TABLE I. ACCURACY (F1 SCORE) OF APP CLASSIFICATION USING IMAGE AND TEXT BASED METHODS

| Category | OCR | Summarization | Object Detection | Text | Ensemble |
|---|---|---|---|---|---|
| Beauty | 0.65 | 0.46 | 0.59 | 0.71 | 0.73 |
| Food | 0.76 | 0.66 | 0.78 | 0.95 | 0.96 |
| Décor | 0.2 | 0.48 | 0.64 | 0.77 | 0.77 |
| Average | 0.64 | 0.56 | 0.7 | 0.84 | 0.86 |

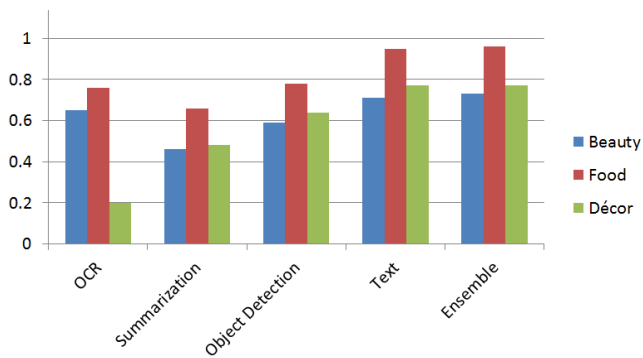

Fig. 3. Accuracy of the app classification for 3 app categories using the image and text based methods.

As we can see, the classification accuracy is best for the ensemble of text and image based methods across all the classes of apps, followed by the text (classification based on textual app description) based method.

In order to determine if the parameters we used (C=1.0 and gamma 0.001) were optimal, we ran a grid search on the text classifier with hyperparameter optimization of C and gamma values for the RBF and linear kernel of SVM. Out of the three classes whose results are plotted in table 1, one class (beauty) performed best with the linear kernel and the other two classes (decor, food) achieved equal performance with the linear and RBF kernels on the validation set. Hence, we determined that the problem was linearly separable.

TABLE II. ACCURACY (F1 SCORE) OF APP CLASSIFICATION USING ENSEMBLE OF IMAGE BASED METHODS WITHOUT AND WITH TEXT CLASSIFIER

| Category | Beauty | Food | Decor | Average |
|---|---|---|---|---|
| Image only | 0.73 | 0.84 | 0.59 | 0.77 |
| Text + image | 0.73 | 0.96 | 0.77 | 0.86 |

In general, the image based methods perform worse than the text method. However, the fact that the accuracy of the ensemble method (combining all 3 image and 1 text based methods and taking a voting) is better shows that using images for app classification helps to improve the accuracy, however slightly, compared to methods where only text is used. One explanation for this is that those apps where the text description was inadequate have been classified better because of the app images. We manually verified that some of the apps (including com.iqdia.megafish, com.foomapp.vendor, io.app.sigdi) in the food category where the text classification was wrong, had short text.

We then analyze the results across categories. We can see that the app classification accuracy for the 'food' category consistently gave better results than the other two categories. This could be because the training data size for the food category was bigger, since there were lots of apps available for the food category in the Google play store compared to other categories. Also, the classification accuracy for 'décor' category was unusually low when the OCR method was used. We examined some of the apps from that category and found that OCR extracted descriptions for the apps from that category had a lot of noise compared to other categories.

We also ran a test to determine how much of the improvement in classification results in the ensemble of methods was attributed to the text method only, and how much improvement was due to the image-based methods. The results are shown in figure 3 across the three categories. Hence as we can see, the text classifier contributed most of the signal for food and décor categories. The F1 score of the standalone text classifier was 0.84 while for the ensemble it was 0.86.

After this, we compare the precision and recall of the three image based methods, for all the apps regardless of category. The results are shown in table 3.

We can see from Table 3 that the precision of the object detection based method is best, while OCR is the best for recall. This is because the RCNN based object detection model correctly detects the objects, giving a higher precision for the overall app classification as well.

On the other hand, the OCR model, which classifies the apps based on extracted text from the app images, correctly detects a slightly higher number of apps. This could be possibly because the extracted text from the images gives more information (more apps have relevant text as part of the images) but much of it is noise, giving a higher recall but lower precision. Identifying the objects gives less information but it is of higher quality, leading to higher precision.

TABLE III. COMPARISION OF THE CLASSIFICATION PRECISION AND RECALL ACCURACY ACROSS IMAGE BASED METHODS

| Method | Precision | Recall | F1-score |
|---|---|---|---|
| OCR | 0.72 | 0.66 | 0.64 |
| Summarization | 0.66 | 0.51 | 0.56 |
| Object Detection | 0.89 | 0.59 | 0.7 |

## V. CONCLUSION AND FUTURE WORK

In this paper we have evaluated multiple methods for app classification using app images. We found that using non text metadata of apps, such as images, can help to better classify the apps in certain cases, compared to purely text based methods. Methods based on extracting text from images using OCR have a higher recall, whereas object detection based methods have a higher precision.

In future, we will analyze other image based methods such as pic2vec, as well as analyze the results for a higher number of app categories.